\documentclass[letterpaper, 10 pt, conference]{ieeeconf}  

\IEEEoverridecommandlockouts
\usepackage{bbold}
\usepackage{amsfonts}
\usepackage{mathtools}

\usepackage{graphicx}
\usepackage{subcaption}

\usepackage{algorithm,algorithmic}

\usepackage[colorinlistoftodos]{todonotes}

\DeclareGraphicsExtensions{.pdf, .png, .jpg}

\newcommand{\states}{\mathcal{S}}
\newcommand{\actions}{\mathcal{A}}
\newcommand{\reals}{\mathbb{R}}
\newcommand{\tr}{^\top}

\newcommand{\x}{\boldsymbol{\phi}}
\newcommand{\w}{\mathbf{w}}
\newcommand{\sr}{\psi}

\newcommand{\I}{I}

\newcommand{\C}{C}
\renewcommand{\c}{c}

\newcommand{\deltav}{\boldsymbol{\delta}}
\newcommand{\avgc}{\bar{\c}}
\newcommand{\avgcb}{\mathbf{\avgc}}
\newcommand{\avggamma}{\bar{\boldsymbol{\gamma}}}
\newcommand{\M}{M}

\usepackage{mdframed}

\title{\LARGE \bf
	Accelerating Learning in Constructive Predictive \\Frameworks with the Successor Representation}

\author{Craig Sherstan$^{1}$, Marlos C. Machado$^{1}$, Patrick M. Pilarski$^{1}$
	\thanks{}
	\thanks{$^{1}$ {\small University of Alberta, Canada \newline \indent \indent \footnotesize \tt{\{sherstan, machado, pilarski\}@ualberta.ca}}}%
}

\begin{document}
	\maketitle
	\thispagestyle{empty}
	\pagestyle{empty}
	
	\begin{abstract}
		
		Here we propose using the \textit{successor representation} (SR) to accelerate learning in a constructive knowledge system based on \textit{general value functions} (GVFs). In real-world settings like robotics for unstructured and dynamic environments, it is infeasible to model all meaningful aspects of a system and its environment by hand due to both complexity and size. Instead, robots must be capable of learning and adapting to changes in their environment and task, incrementally constructing models from their own experience. GVFs, taken from the field of reinforcement learning (RL), are a way of modeling the world as predictive questions. One approach to such models proposes a massive network of interconnected and interdependent GVFs, which are incrementally added over time. It is reasonable to expect that new, incrementally added predictions can be learned more swiftly if the learning process leverages knowledge gained from past experience. The SR provides such a means of separating the dynamics of the world from the prediction targets and thus capturing regularities that can be reused across multiple GVFs. As a primary contribution of this work, we show that using SR-based predictions can improve sample efficiency and learning speed in a continual learning setting where new predictions are incrementally added and learned over time. We analyze our approach in a grid-world and then demonstrate its potential on data from a physical robot arm.
		
	\end{abstract}
	
\section{INTRODUCTION}
	
	A long standing goal in the pursuit of artificial general intelligence is that of knowledge construction---incrementally modeling and explaining the world and the agent's interaction with it directly from the agent's own experience. This is particularly important in fields such as continual learning~\cite{Ring1994} and developmental robotics~\cite{Oudeyer07}, where we expect agents to be capable of learning, dynamically and incrementally, to interact and succeed in complex environments. 
	
	One proposed approach for representing such world models is a collection of \textit{general value functions} (GVFs)~\cite{Sutton2011}, which models the world as a set of predictive questions each defined by a policy of interest, a target signal of interest under that policy, and a timescale (discounting schedule) for accumulating the signal of interest. For example, a GVF on a mobile robot could pose the question ``How much current will my wheels consume over the next second if I drive straight forward?''
    
    GVF questions are typically answered using temporal-difference (TD) methods~\cite{Sutton1988} from the field of reinforcement learning (RL)~\cite{Sutton1998}. 
	A learned GVF approximates the expected future value of a signal of interest, directly representing the relationship between the environment, policy, timescale, and target signal as the output of a single predictive unit.
	
	Nevertheless, despite the success RL algorithms have achieved recently (e.g., \cite{Mnih2015,Silver2017}), methods for answering multiple predictive questions from a single stream of experience (critical in a robotic setting) are known to exhibit sample inefficiency. In our setting of interest, where multitudes of GVFs are learned in an incremental, sample by sample way, this problem is multiplied. Ultimately, the faster an agent can learn to approximate a new GVF, the better.
	
	In this paper we show how one can accelerate learning in a constructive knowledge system based on GVFs by sharing the environment dynamics across the different predictors.
	This is done with the successor representation (SR)~\cite{Dayan1993}, which allows us to learn the world dynamics under a policy independently of any signal being predicted.
	
	We empirically demonstrate the effectiveness of our approach on both a tabular representation and on a robot arm which uses function approximation. We evaluate our algorithm in the continual learning setting where it is not possible to specify all GVFs {\em a priori}, but rather GVFs are added incrementally during the course of learning. As a key result, we show that using a learned SR enables an agent to learn newly added GVFs faster than when learning the same GVFs in the standard fashion without the use of the SR.
	
	
\section{BACKGROUND}
	
	We consider an agent interacting with the environment sequentially. We use the standard notation in the reinforcement learning (RL) literature~\cite{Sutton1998},
	modeling the problem as a Markov Decision Process. Starting from state $S_0\in\states$, at each timestep the agent chooses an action, $A_t\in\actions$, according to the policy distribution $\pi:\states\times\actions\rightarrow[0,1]$, and transitions to state $S_{t+1}\in\states$ according to the probability transition $p(\cdot | S_t,A_t)$. For each transition $S_t\xrightarrow{A_t}S_{t+1}$ the agent receives a reward, $R_t$, from the reward function $R(S_t,A_t,S_{t+1})\in\reals$. 
	
	In this paper we focus on the prediction problem in RL, in which the agent's goal is to predict the value of a signal from its current state (e.g., the cumulative sum of future rewards). Note that throughout this paper we use upper case letters to indicate random~variables.
	
	\subsection{General Value Functions (GVFs)}
	
	The most common prediction made in RL is about the expected return. The return is defined to be the sum of future discounted rewards under policy $\pi$ starting from state~$s$. Formally, $G_t = \sum_{t=0}^T \gamma^t R_t$, with $\gamma \in [0,1]$ being the discount factor and $T$ being the final timestep, where $T=\infty$ denotes a continuing task. The function encoding the prediction about the return is known as the value function $v_{\pi}(s) = \mathbb{E}_\pi[G_t | S_0 = s]$.
	
	GVFs~\cite{Sutton2011} extend the notion of predictions to different signals in the environment. This is done by replacing the reward signal, $R_t$, by any other target signal, which we refer to as the cumulant, $\C_t$, and by allowing a state-dependent discounting function, $\gamma_t=\gamma(S_t)$, instead of using a fixed discounting factor. The general value of state $s$ under policy $\pi$ is defined as:
	\begin{align}\label{eq:gvfs}
	v_{\pi:\gamma}(s) &=\bar{\c}_{\pi} + \bar{\gamma}_{\pi}\sum_{s'}p_{\pi}(s'|s)v_{\pi}(s'),
	\end{align}
	where $\bar{\c}_\pi$ is the average cumulant from state $s$, $\bar{\gamma}_{\pi}$ is the average $\gamma$ from state $s$, and $p_{\pi}(s'|s)$ is the probability of transitioning from state $s$ to $s'$ under policy $\pi$. This can also be written in matrix form: $\textbf{v}_{\pi:\gamma}=\avgcb_\pi + \avggamma_\pi\odot P_\pi \textbf{v}_{\pi:\gamma}$, where $\odot$ denotes element-wise multiplication.. Such an equation, when solved, gives us
	\begin{align}\label{eq:sol_gvfs}
	\textbf{v}_{\pi:\gamma}=(\I-\avggamma_\pi\odot P_\pi)^{-1}\avgcb_\pi,
	\end{align}
	where $\I$ is the identity matrix, $\textbf{v},\avgcb,\avggamma\in\reals^{|\states|\times1}$, and $P_\pi \in [0,1]^{|\states|\times|\states|}$ is a probability matrix such that $[P_{\pi}]_{ij}=\mathbb{E}[p_\pi(S_{t+1}=s_j|S_t=s_i)]$.
	
\subsection{The Successor Representation (SR)}
	
	The successor representation~\cite{Dayan1993} was initially proposed as a representation capable of capturing state similarity in terms of time. It is formally defined, for a fixed $\gamma < 1$, as:
	$$\psi_\pi(s,s') = \mathbb{E}_\pi \Big[\sum_{t=0}^\infty \gamma^t \mathbb{1}_{\{S_t = s'\}} | S_0 = s\Big].$$
	In words, the SR encodes the expected number of times the agent will visit a particular state when the sum is discounted by $\gamma$ over time\footnote{Note that Dayan describes the SR as predicting future state visitation from time $t$ onward. This is non-standard in RL as we typically describe the return as predicting the signal from $t+1$ onward.}. It can be re-written in matrix form as
	\begin{eqnarray}\label{eq:sr_solution}
	\Psi_\pi = \sum_{t=0}^\infty (\gamma P_\pi)^t = (I - \gamma P_\pi)^{-1}.
	\end{eqnarray}
	
	Importantly, the SR can be easily computed, incrementally, by standard RL algorithms such as TD learning~\cite{Sutton1988}, since its primary modification is to replace the reward signal by a state visitation counter. Nevertheless, despite its simplicity, the SR holds important properties we leverage in this paper. 
	
	The SR, in the limit, for a constant $\gamma$ (see Eq.~\ref{eq:sr_solution}), corresponds to the first factor of the solution in Eq.~\ref{eq:sol_gvfs}. Thus, the SR can be seen as encoding the dynamics of the Markov chain induced by the policy $\pi$ and by the environment's transition probability $p$. If the agent has access to the SR, it can accurately predict the (discounted) accumulated value of any signal, from any state, by simply learning the expected \emph{immediate} value, $\avgcb_\pi$, of that signal in each state. On the other hand, if the agent does not use the SR, the agent must also deal with the problem of credit assignment, having to look at $n$-step returns to control for \emph{delayed} consequences. Importantly, the dynamics encoded by the SR are the same for all signals learned under the same policy and discount function. This factorization of the solution is the main property we use in our work, described in the next section.
	
\section{METHODS}
    
    As aforementioned, we are interested in the problem of knowledge acquisition in the continual learning setting~\cite{Ring1994}, where knowledge is encoded as predictive questions (GVFs). In this setting it is not possible to specify all GVFs ahead of time. Instead, GVFs must be added incrementally by some, as yet unknown, mechanism. The standard approach would be to learn each newly added prediction from scratch. In~this section we discuss how we can use the SR to accelerate learning by taking advantage of the factorization shown in Eq.~\ref{eq:sol_gvfs}. Our method leverages the fact that the~SR~is~independent of the target signal being predicted, learning the SR separately and re-using it when learning to predict new signals.
	
	In the previous section, for clarity, we discussed the main concepts in the tabular case. In real world applications, where the state space is too large, assuming states can be uniquely identified is not often feasible. Instead, we generally represent states as a set of features $\x(s) \in \mathbb{R}^d$ where $d \ll |\states|$. Because both $\sr$ and $\avgc$ are a function of feature vector $\x(S)$, they can easily be represented using function approximation and learned using TD algorithms. In order to present a more general version of our algorithm, we introduce it here using the function approximation notation. 
	
	The first step in our algorithm is to compute the one-step average cumulant, which we do with TD error:
	\begin{align}
	\delta_t=C_{t+1} - \avgc\big(\x(S_t)\big).
	\label{eq:td_c}
	\end{align}
	If we use linear function approximation to estimate $\avgc$ then $\avgc\big(\x(s)\big)=\x(s)\tr\w$. The TD error for $\sr$ is given as (note that $\deltav$ is a vector of length $n$) 
	\begin{align}
	\deltav_t=\x(S_{t}) + \gamma_{t+1} \sr\big(\x(S_{t+1})\big) - \sr\big(\x(S_t)\big).
	\label{eq:td_sr}
	\end{align}
	This generalization of the SR to the function approximation case is known as successor features~\cite{Barreto2017}.
	
	If we use linear function approximation to estimate $\sr$ then $\sr\big(\x(s)\big)=\M\tr\x(s)$, where $\M \in \mathbb{R}^{d \times d}$. Using the usual semi-gradient method,\footnote{In semi-gradient methods, the effect of $\M$ on the prediction target $\x(S_{t+1}) + \gamma \sr\big(\x(S_{t+1})\big)$ is ignored when computing the gradient.} often used with TD, we derive a TD(0) stochastic gradient descent update as:
	\begin{align*}
	\M_{t+1}&=\M_t - \frac{1}{2}\alpha\nabla_{\M}(\deltav_t^2)\\
	&=\M_t + \alpha\nabla_{\M}\big(\sr(S_t)\big)\otimes\deltav_t\\
	&=\M_t + \alpha \x(S_t) \otimes \deltav_t,
	\end{align*}
	where $\nabla_{\M}$ is the gradient with respect to $\M$ and $\otimes$ is the outer product. Based on this derivation, as well as Eq.~\ref{eq:td_c}~and~\ref{eq:td_sr}, we obtain Algorithm~\ref{alg}. Note that the last two lines of Algorithm~\ref{alg}, which update the SR for the state $S'$, are only required for the episodic case.
	
	\floatname{algorithm}{Algorithm}
	\vspace{0pt}  
	\begin{algorithm}[t]
		\caption{\ \ GVF prediction with the SR}\label{alg:general}
		\begin{algorithmic}
			\STATE \textbf{Input:} Feature representation $\phi$, policy $\pi$, discount function $\gamma$, and step-sizes $\alpha_{\footnotesize C}$, $\alpha_{\mbox{\footnotesize SR}}$
			\STATE \textbf{Output:} Matrix $M$ and vectors $\avgc_i$ as predictors of $C_i$ \vspace{0.2cm}
			\STATE Initialize $\w$ and $M$ arbitrarily \vspace{0.1cm}
			\WHILE {$S'$ is not terminal}
			\STATE Observe state $S$, take action $A$ selected according to $\pi(S)$, and observe a next state $S'$ and the cumulants $C_i$
			\STATE $\delta_{\mbox{\footnotesize SR}}=\phi(S) + \gamma(S') M^\top \phi(S') - M^\top \phi(S)$ \vspace{0.1cm}
			\STATE $M \leftarrow M + \alpha_{\mbox{\footnotesize SR}} \phi(S) \otimes \delta_{\mbox{\footnotesize SR}}$
			
			\FOR {\textbf{each} cumulant ${C}_i$} \vspace{0.1cm}
			\STATE $\delta_{\footnotesize C_i} = C_i - \phi(S)^\top \w_i$ \vspace{0.1cm}
			\STATE $\w_i \leftarrow \w_i + \alpha_{\footnotesize C_i} \phi(S) \delta_{\footnotesize C_i}$ \vspace{0.25cm}
			
			
			\ENDFOR
			\ENDWHILE
			
			\STATE $\delta_{\mbox{\footnotesize SR}}=\phi(S') - M^\top \phi(S')$ \vspace{0.1cm}
			\STATE $M \leftarrow M + \alpha_{\mbox{\footnotesize SR}} \phi(S') \otimes \delta_{\mbox{\footnotesize SR}}$
		\end{algorithmic}
		\label{alg}
	\end{algorithm}
	
	This algorithm allows us to predict the cumulant $C_i$, in state $S$, using the current estimate of the matrix $M$ and the weights $\w_i$. We can then obtain the final prediction by simply computing $\sr\big(\phi(S)\big)^{\tr}\w =\big(\M^\top\x(s)\big)^{\tr}\w = \x(s)^{\tr}\M\w.$
	
	This algorithm accelerates learning because, generally, learning to estimate $\avgc$ is faster than learning the GVF directly. This is exactly what our algorithm does. When predicting a new signal, it starts with its current estimate for the SR, $\Psi$. At the end, the multiplication $\Psi \avgc$  is simply a weighted average of the one-step predictions across all states, weighted by the likelihood they will be visited. We provide empirical evidence supporting this claim in the next sections.
	
\section{EVALUATION IN DAYAN'S GRID WORLD}
	
\begin{figure*}[t!]
		\begin{center}
			\begin{subfigure}[b]{0.3\textwidth}
				\raisebox{6mm}{\centerline{\includegraphics[width=\textwidth]{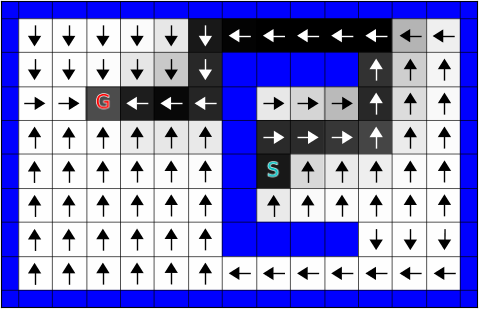}}}
				\caption{}
				\label{fig:dayan_sr}
			\end{subfigure}
			~
			\begin{subfigure}[b]{0.3\textwidth}
				\centerline{\includegraphics[width=\textwidth]{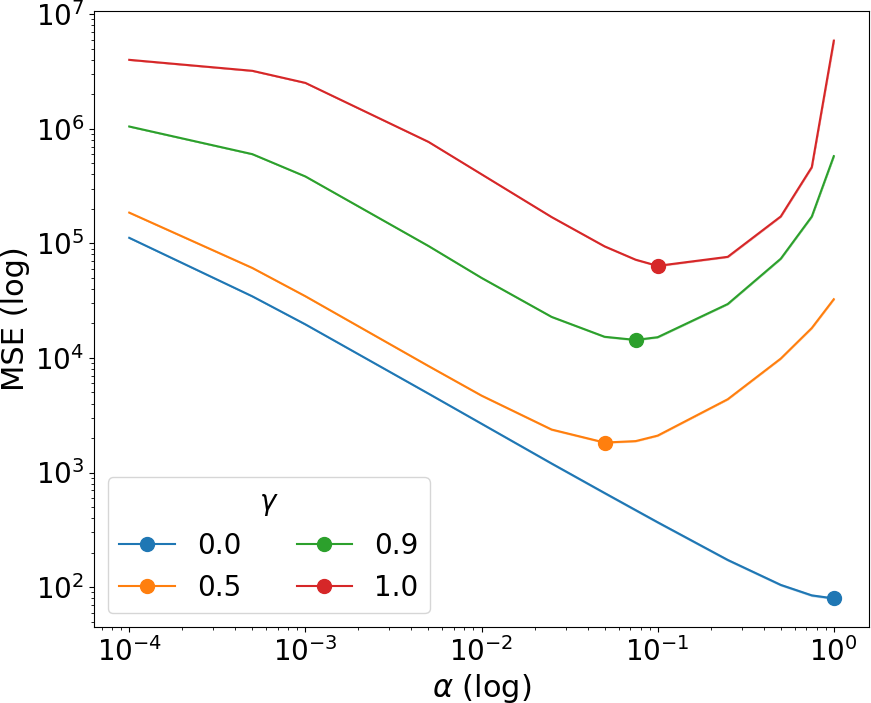}}
				\caption{}
				\label{fig:dayan_sr_alphas_sr}
			\end{subfigure}
			~
			\begin{subfigure}[b]{0.3\textwidth}
				\centerline{\includegraphics[width=\textwidth]{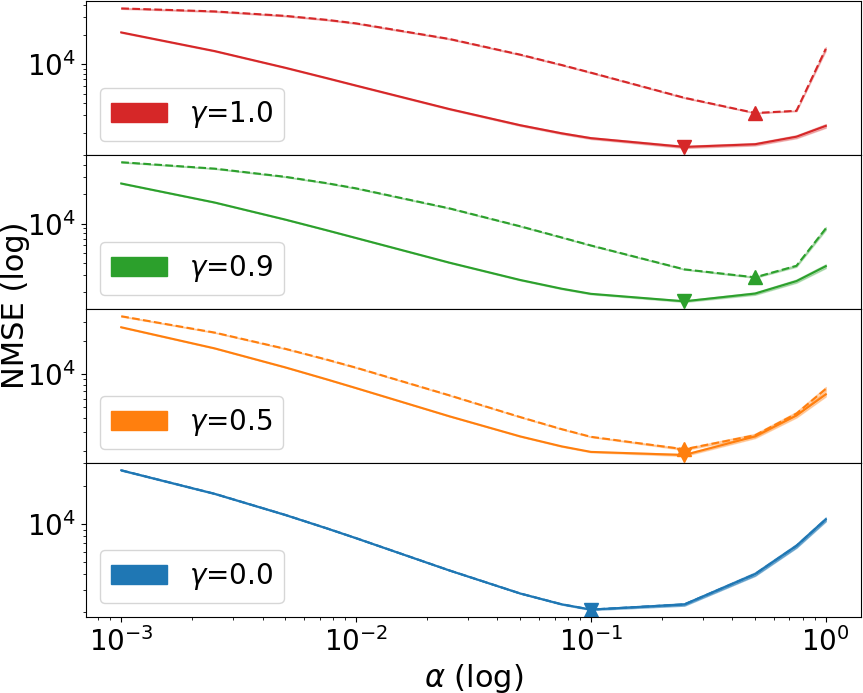}}
				\caption{}
				\label{fig:dayan_p_alphas}
			\end{subfigure}		
		\end{center}
        		\vspace{-0.6em}
		\caption{\textbf{a)} Dayan's grid world. Arrows indicate the hand-coded policy (from the start state the policy is to go up). Black squares indicate the SR prediction given from the starting state, $S$, for $\gamma=1.0$; the darker the square the higher the expected visitation. Notice the graying around the central path caused by the $\epsilon$-greedy action selection. \textbf{b)} MSE of the SR as a function of step-size $\alpha$ for different values of $\gamma$. Lowest error is indicated by the markers. \textbf{c)} A comparison of the NMSE of the direct (dashed lines) and SR-based (solid lines) predictions as a function of fixed step-size $\alpha$ and discount factor $\gamma$ summed across all signals. Lowest cumulative error is indicated by up arrows (direct predictions) and down arrows (SR-based predictions). Note, that although difficult to see, confidence intervals of 95\% are included in both b and c.}
		\vspace{-1em}
	\end{figure*}

    We first evaluated our algorithm in a tabular grid world. Its simplicity allowed us to analyze our method more thoroughly since we were not bounded by the speed and complexity of physical robots. The grid world we used was inspired by Dayan's~\cite{Dayan1993} (see Figure~\ref{fig:dayan_sr}). Four actions are available in this environment: up, down, left, right. Taking an action into~a wall (blue) results in no change in position. Transitions are deterministic, i.e., a move in any direction moves the agent to the next cell in the given direction, except when moving into a wall. For each episode the agent spawns at location \textbf{S} and the episode terminates when the agent reaches the goal~\textbf{G}.  

	We generated fifty different signals for the agent to predict. They were generated randomly from a collection of primitives enumerated in Table~\ref{table:sigs}. They are composed of two different primitives, one for each axis, like so: $(sig_x(x + offset_x) + bias_x)*(sig_y(y + offset_y) + bias_y)$. The bias and offset were drawn from $[-2.0,2.0)$ and $[0, 10)$, respectively. Offset and bias were not applied to either the unit or shortest path primitives. Further, the shortest path primitive was not combined with a second signal, but was used on its own. Gaussian noise with a standard deviation of 0.3 was applied on top of each signal. The shortest path signal is inspired by a common reward function used in RL where each transition has a cost (a negative reward) meant to push the agent to completing a task in a timely manner and reaching the goal produces a positive reward signal.

Our agent selects actions using $\epsilon$-greedy action selection where, at each timestep,  with probability $1-\epsilon$, it uses the action specified by a hand-coded policy (see Figure~\ref{fig:dayan_sr}), and otherwise chooses randomly from all four actions ($\epsilon=0.3$ in our experiments). A tabular representation is used with each grid cell uniquely represented by a one-hot encoding.

		\begin{table}[t]
			\caption{Signal Primitives} \label{table:sigs}
            \vspace{-1em}
			\begin{center}
				\begin{tabular}{|p{1.7cm}|p{5cm}|}
					\hline
					\textbf{Primitive} & \textbf{Parameters} \\
					\hline
					Fixed value & value~$\in[-2.0,2.0)$ \\
					\hline
					Square wave & period~$\in[2,40)$, invert~$\in[True,False]$\\
					\hline
					Sin wave & period~$\in[2,40)$\\
					\hline
                    Random binary & Fixed random binary string generated over the length of an axis.\\
					\hline
					Random float & Fixed random floats $\in[0,1.0)$  generated over the length of an axis.\\
					\hline
					Unit & Fixed value of 1.0\\
					\hline
					Shortest path & transition\_cost~$\in [-10.0,-1.0)$,\\ 
                    & goal\_reward $\in[1.0,10.0)$\\
					\hline			
				\end{tabular}
			\end{center}
			\vspace{-2em}
		\end{table}

	In this set of experiments we can compute the ground-truth predictors for the SR and the signal predictors. This is done by taking the average return observed from each state. The SR reference was averaged over 30,000 episodes and the signal predictor references were averaged over 10,000 episodes. Each episode started at the start state and followed the $\epsilon$-greedy policy already described.
	
    We first evaluated the predictive performance of our SR learning algorithm with respect to the step-size for a variety of $\gamma$ values. We report the average over 30 trials of 10,000 episodes. We initialize the SR weights to 0.0. The squared Euclidean distance was calculated between the predicted SR and the reference SR for each timestep. These values were summed over the run and the average was taken across the runs. These averages are shown in Figure~\ref{fig:dayan_sr_alphas_sr}. 

	Using the results in Figure~\ref{fig:dayan_sr_alphas_sr} we 
    evaluated the performance of the two signal prediction approaches by sweeping across step-sizes for various $\gamma$. For each experimental run, learning of a new signal was enabled incrementally every 50 episodes. This produced runs with a total length of 2,500 episodes where the first pair of GVFs added (the direct and one-step predictors) were trained for 2,500 episodes and the last added GVFs were trained for 50 episodes. Further, for each run, the order in which the signals were added was randomized. Thirty runs were performed. The weights of the predictors and of the SR were initialized to 0.0. Notice that the SR was being learned at the same time as the direct and one-step predictions.

For each run a cumulative MSE for each signal $i$ was calculated according to Eq.~\ref{eq:dayan_mse}. This equation computes the total squared error between the predictor's estimate, $V$ and the reference predictor's estimate, $V^*$. For each episode, $E$, the error of the current and previous episodes is averaged. Then, for each signal, the maximum error for a given $\gamma$, either in the direct or SR-based predictions, is found and used to normalize the errors in the signal across that particular value of $\gamma$ (see Eq.~\ref{eq:dayan_nmse}). In this way we attempt to treat the error of each signal equally. If this is not done the errors of large magnitude signals dominate the results. These normalized values are then summed across the signals and the averages across all 30 runs are plotted in Figure~\ref{fig:dayan_p_alphas}.

\begin{align}
MSE_{i} =&\frac{1}{E}\sum_{e_0}^E\sum_t^T(V_{i:t} - V_{i:t}^*)^2 \label{eq:dayan_mse} \\
NMSE_{i:\alpha:\gamma}=&\frac{MSE_{i:\alpha:\gamma}}{\max_{\alpha}(MSE_{i:\gamma:direct},MSE_{i:\gamma:SR })} \label{eq:dayan_nmse} 
\end{align}

The advantage of the SR-based method is clear as $\gamma$ increases. This is to be expected since for $\gamma=0.0$ both methods are making one-step predictions. In the  experiment of Figure~\ref{fig:dayan_p_alphas}, the SR performs better in the vast majority of the signals, as shown in Table~\ref{table:better_signals}. For all step-sizes not listed the SR-based method was better on all signals. Analysis of these cases where the direct method did better reveal that some of the target signals have very small magnitudes, suggesting the SR-based approach may be more susceptible to signal-to-noise ratio. Further analysis remains to be done.

\begin{table}[b]
	\vspace{-1em}
	\caption{Signal performance for $\gamma=0.9$ of Figure~\ref{fig:dayan_p_alphas}.} \label{table:better_signals}
    \vspace{-1em}
	\begin{center}
		\begin{tabular}{|l|c|c|}
			\hline
			$\alpha$ & \textbf{Direct Better} & \textbf{SR-Based Better} \\
			\hline
			0.25 & 3 & 47 \\
			\hline
			0.5 & 5 & 45 \\
			\hline
			0.75 & 4 & 46 \\
			\hline
			1.0 & 1 & 49\\
			\hline
		\end{tabular}
	\end{center}
	\vspace{-1em}
\end{table}

\begin{figure}
		\begin{center}
			\centerline{\includegraphics[height=2.0in]{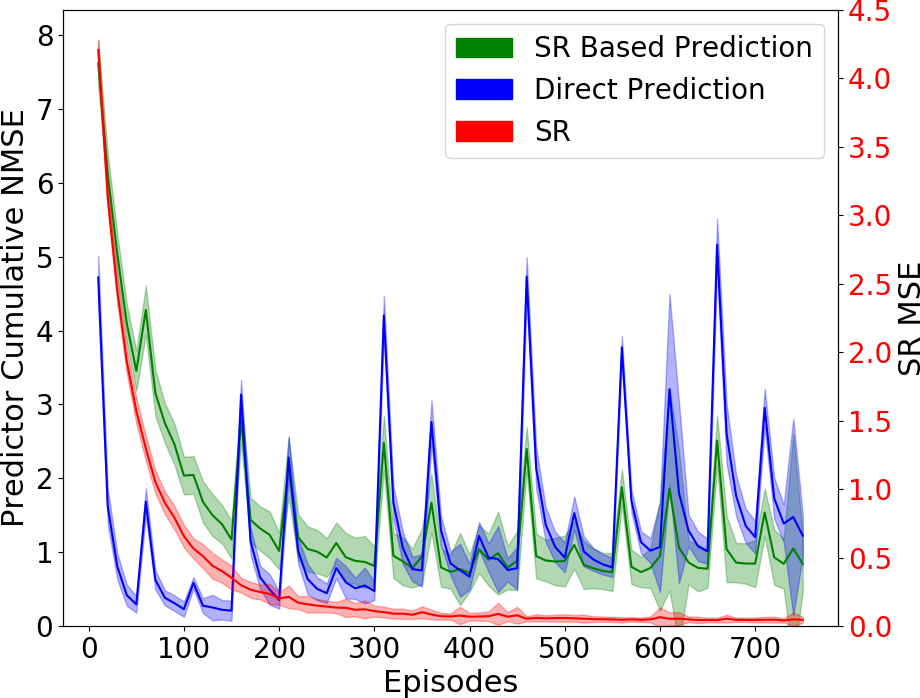}}
			\caption{All predictors learn from scratch with new predictors added in every 50 episodes. As the SR error (red, right axis) goes low the SR-based predictors (green) are able to learn faster than their direct (blue) counterparts. Shading indicates a 95\% confidence interval.}
			\label{fig:dayan_wave}
            \vspace{-3em}
		\end{center}
	\end{figure}

Finally, we analyzed how the prediction error of our systems evolve with time. This is demonstrated in Figure~\ref{fig:dayan_wave} where we selected the best step-sizes for $\gamma=0.9$ and plotted the performance over time across 30 different runs. In this case the order of the signals remained fixed so that sensible averages could be plotted for each signal. Signal performance was normalized as before and summed across all active signals. As expected, we clearly see that the SR-based predictions (green) start with much higher error than the direct (blue), but as the error of the SR (red) drops low the newly added SR-based predictors are able to learn quicker, with less peak and overall error than the direct predictor.

In the continual learning setting we never have the opportunity to tune for optimal step-sizes as we did in our evaluation. Practically, fixed step-sizes are used for many robotics settings in RL, but, in order to ensure stable learning, small step-sizes are chosen. As we saw in Figure~\ref{fig:dayan_p_alphas}, the advantage of using the SR-based predictions is enhanced with smaller step-sizes. Ideally, however, we would imagine that a fully developed system would use some method of adapting step-sizes, such as ADADELTA~\cite{Zeiler2012}.

\section{EVALUATION ON A ROBOT ARM}

\begin{figure}[t!]
		\begin{center}
			\centerline{\includegraphics[height=1.8in]{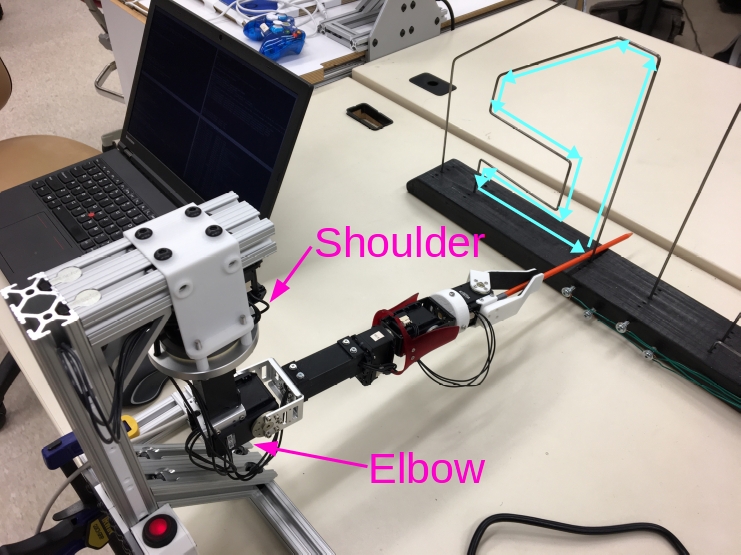}}
			\caption{The user controls the robot arm using a joystick to trace the inside of the wire maze in a counter-clockwise direction. Circuit path shown in blue.}
			\label{fig:robot_maze}
            \vspace{-3em}
		\end{center}
	\end{figure}

	Tabular settings like Dayan's grid world are useful for enabling analysis and providing insight into the behavior of our method. However, our goal is to accelerate learning on a real robot where states are not fully observed and cannot be represented exactly; instead we must use function approximation. Here we demonstrate our approach using a robot arm and learning sensorimotor predictions with respect to a human-generated policy. In our task a user controls a robot arm 
    via joystick to trace a counter-clockwise circuit through the inside of the wire maze (see Figure~\ref{fig:robot_maze}) with a rod held in the robot's gripper. The user performed this task for approximately 12 minutes completing around 50 circuits.
    
	In this experiment we used six different prediction targets: the current, position and speed of both the shoulder rotation and elbow flexion joints. A new predictor was activated every 2,000 timesteps (Note that the robot reports sensor updates at 30 timesteps/s). For this demonstration a discount factor of $\gamma=0.95$ was used. Four signals were used as input to our function approximator: the current position and a decaying trace of the position for the shoulder and elbow joints. The decaying trace for joint $j$ was calculated as $tr_{j:t+1}=0.8*tr_{j:t} + 0.2*pos_{j:t+1}$. These inputs were normalized over the joint ranges observed in the experiment and passed into a 4-dimensional tilecoding~\cite{Sutton1998} with 100 tilings of width 1.0 and a total memory size of 2048. Additionally a bias unit was added, resulting in a binary feature vector of length 2049 with a maximum of 101 active features on each timestep (hashing collisions can reduce this number). We use a decaying step-size for all the predictors where the step-size starts at 0.1 and decays linearly to zero over the entire dataset. At each timestep this step-size is further divided by the number of active features in $\phi(S_{t})$. Finally, for each predictor, this step-size is offset such that the step-size starts at 0.1 when it is first activated and decays at the same rate as all the other predictors.
	
To compare the prediction error we compute a running MSE for each signal according to Eq.~\ref{eq:robot_mse}, where at each timestep $t$ the sum is taken over all previous timesteps. Unlike the previous tabular domain, we do not have the ideal estimator to compare against and instead compare the predictions, $V$, against the actual return, $G$. In order to treat each signal equally we further normalize these errors according to Eq.~\ref{eq:robot_nmse}. Note that the NMSE allows us to compare the predictions of a single signal between the two methods, but does not tell us how accurate the predictions are nor does it allow comparison between signals.
\begin{align}
	MSE_{i:t} =& \frac{1}{T}\sum_{t=t_0}^T(V_{i:t} - G_{i:t})^2 \label{eq:robot_mse}
\end{align}

\begin{align}
	NMSE_{i}=&\frac{MSE_{i}}{\max(MSE_{i:direct},MSE_{i:SR})} \label{eq:robot_nmse}     
\end{align}
	
\begin{figure}[t!]
		\begin{center}
          \centerline{\includegraphics[height=1.8in]{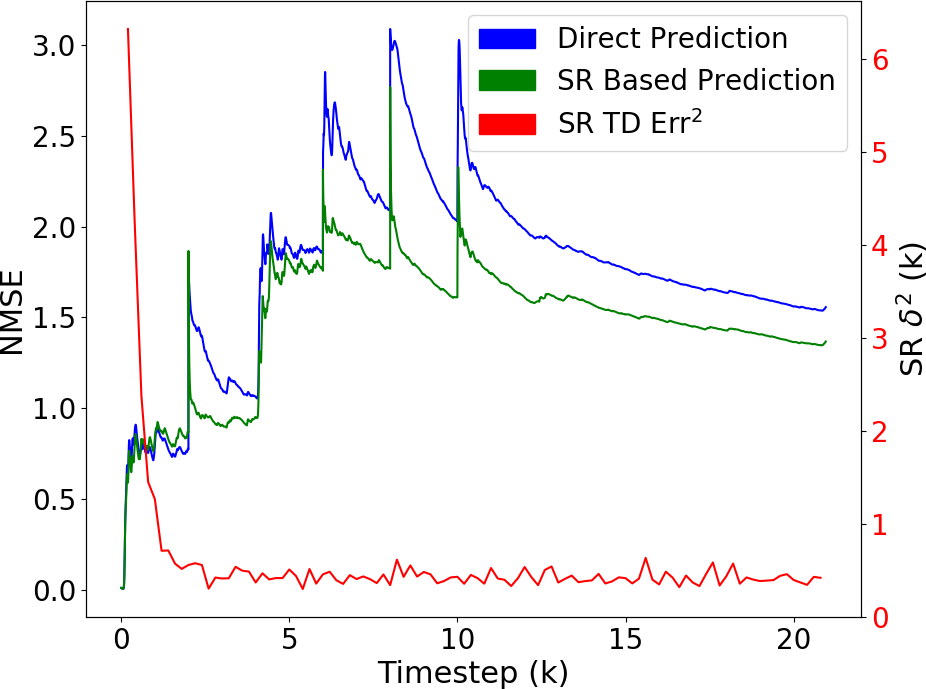}}
          \caption{A 12 minute run tracing the maze circuit. A new predictor is added every 2,000 timesteps. NMSE errors are summed across all predictors.}
          \label{fig:robot_wave}
          \vspace{-2em}
		\end{center}
	\end{figure}

\begin{figure}[t!]
		\begin{center}
          \centerline{\includegraphics[width=2.3in]{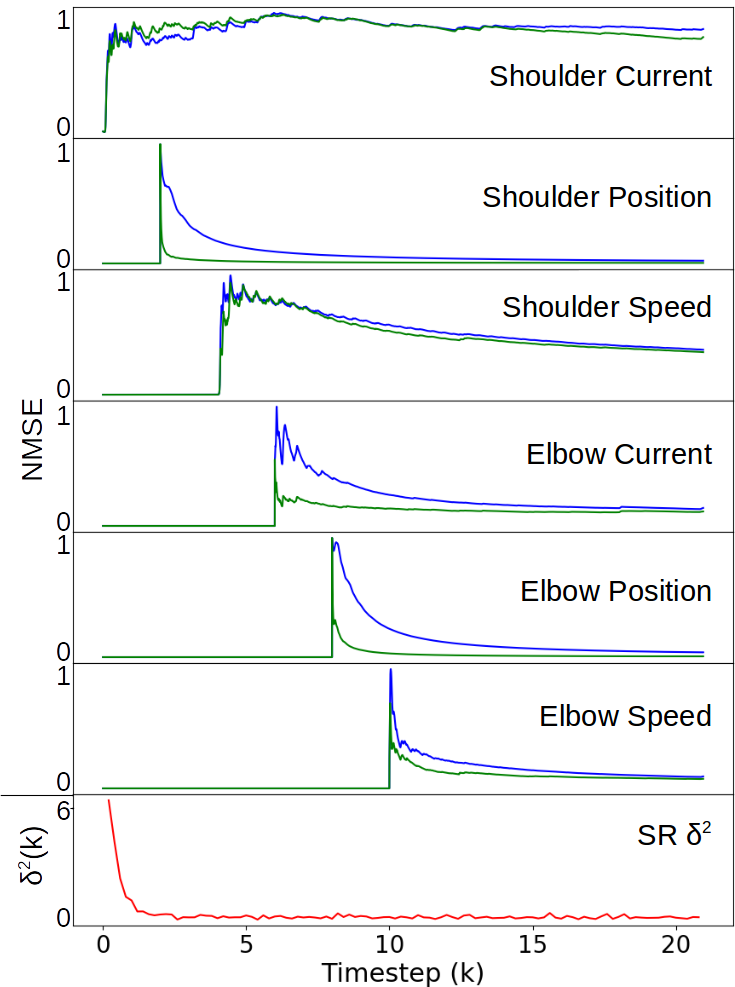}}
          \caption{The same results as Figure~\ref{fig:robot_wave}, but with the NMSE for the individual predictors. Each is normalized from 0 to 1.}
          \label{fig:robot_stacked_wave}
          \vspace{-2.5em}
		\end{center}        
	\end{figure}

Figures~\ref{fig:robot_wave} and \ref{fig:robot_stacked_wave} show a single run, approximately 12 minutes in length. A single ordering of the predictors was used. Figure~\ref{fig:robot_wave} shows the error across all predictors while Figure~\ref{fig:robot_stacked_wave} separates out each predictor. Here we see a clear advantage to using the SR-based predictions for most of the signals. Unlike the previous tabular results, there is little difference on the performance of the first predictor (shoulder current), even while the SR is being learned. To investigate, we ran experiments where each signal was learned from the beginning of the run. We observed that performance was rarely worse and sometimes even better when using the SR-based method. This suggests the SR-based approach is more robust than expected, but further experimentation is needed.
	
	\section{FURTHER ADVANTAGES WHEN SCALING}
	
	While this paper analyzed single policies and discount functions, this is not the setting in which the GVF framework is proposed to be used. Rather, it is imagined that massive numbers of GVFS over many policies and timescales will be used represent complex models of the world \cite{Sutton2011,Modayil2014a}. In this setting we note that using SR-based predictions can offer additional benefits, allowing the robot to do more with less. Consider, for a single policy $\pi$, a collection of SRs learned for $f$ discount functions and $h$ one-step predictors. We can then represent $fh$ predictions using $f + h$ predictors. A first advantage is that far fewer GVFs need to be updated on each timestep, saving computational costs. As a second benefit, there is potential to reduce the number of weights used by the system. For example, consider learning in a tabular setting, with $|\states|$ states, using linear estimators. For $fh$ predictions the number of weights needed is : $|w|_{direct}=fh|\states|$, $|w|_{SR-based}=f|\states|^2 + h|\states|$. It can be shown that for a fixed $f$ and $\states$ the total number of weights used by the direct prediction approach is greater when $h>\frac{f|\states|}{f-1}$.
    

\section{RELATED WORK}
	
	The idea of the SR was originally introduced as a function approximation method \cite{Dayan1993}. However, it has recently been applied to other settings. It~has been used, for instance, in transfer learning problems allowing agents to generalize better across similar but different tasks~\cite{Barreto2017,Kulkarni2016}, and to define intrinsic rewards in option discovery algorithms~\cite{Machado2018}.
	
	GVFs were originally proposed as a method for building an agent's overall knowledge in a modular way~\cite{Sutton2011}. To date they have primarily been used with hand-coded fixed policies~\cite{Modayil2014a}. The UNREAL agent~\cite{Jaderberg2017} is a powerful demonstration of the usefulness of multiple predictions; pre-defined auxiliary tasks, which can be viewed as GVFs, are shown to accelerate and improve the robustness of learning.
    
	
	Finally, the idea closest to this work is the concept of Universal Value Functions (UVFAs)~\cite{Schaul2015}. UVFAs are, as GVFs, a generalization of value functions. However, instead of generalizing them to multiple predictors (and discount factors) they generalize value functions over goals in a parametrized way. We believe our result and the idea of UVFAs are complementary, and could in fact be eventually combined in a future work.

	\section{CONCLUSIONS}
	
	In this paper we showed how the successor representation (SR), although originally introduced for another purpose, can be used to accelerate learning in a continual learning setting in which a robot incrementally constructs models of its world as a collection of predictions known as general value functions (GVFs). The SR enables a given prediction to be modularized into two components, one representing the dynamics of the environment (the SR) and the other representing the target signal (one-step signal prediction). This allows a robot to reuse its existing knowledge when adding a new prediction target, speeding up learning of the new predictor. We demonstrated this behaviour in both a tabular grid world and on a robot arm. These results suggest an effective method for improving the learning rate and sample efficiency for robots learning in the real world.
	
	There are several clear opportunities for further research on this topic. The first is to provide greater understanding into why, for a given fixed step-size, some (few) signals are better predicted directly rather than through the SR. Further, the work in using the SR with function approximation is preliminary and more insight can yet be gained in this setting. Another opportunity for research is to explore using SR-based predictions with state-dependent discount functions. Finally, we suggest that SR-based predictions with deep feature learning \cite{Kulkarni2016,Machado2018} and an incrementally constructed architecture would be a very powerful tool to support continual or developmental learning in robotic domains with widespread real-world applications.

	\addtolength{\textheight}{-12cm}   
	
%
	
	\bibliography{main}
	\bibliographystyle{IEEEtran}
	
	\clearpage

\end{document}